\documentclass[11pt,a4paper]{article}
\usepackage[hyperref]{emnlp-ijcnlp-2019}
\usepackage{times}
\usepackage{latexsym}
\usepackage{amsfonts,amssymb}
\usepackage{amsmath}
\usepackage{xspace}
\usepackage{amsfonts}
\usepackage{url}
\usepackage{graphicx}
\usepackage{subfigure}
\usepackage{float}
\usepackage{multirow}
\usepackage{booktabs}

\aclfinalcopy % Uncomment this line for the final submission
\newcommand{\argmax}{\operatornamewithlimits{argmax}}

\newcommand{\sos}{\mbox{\texttt{<s>}}}
\newcommand{\eos}{\mbox{\texttt{</s>}}}

\title{Latent-Variable Generative Models for Data-Efficient Text Classification}
\author{Xiaoan Ding$^1$ \ \ \  Kevin Gimpel$^2$ \\[1ex]
$^1$University of Chicago, Chicago, IL, 60637, USA\\
$^2$Toyota Technological Institute at Chicago, Chicago, IL, 60637, USA\\
[1ex]
{
{\tt xiaoanding@uchicago.edu}, \ \ 
{\tt kgimpel@ttic.edu}
}\\}

\date{}

\begin{document}
\maketitle
\begin{abstract}

Generative classifiers offer potential advantages over their discriminative counterparts, namely in the areas of data efficiency, robustness to data shift and adversarial examples, and zero-shot learning \citep{ng2002discriminative, yogatama2017generative, lewis2018generative}. 
In this paper, we improve generative text classifiers by introducing discrete latent variables into the generative story, and explore several graphical model configurations. We parameterize the distributions using standard neural architectures used in conditional language modeling and perform learning by directly maximizing the log marginal likelihood via gradient-based optimization, which avoids the need to do expectation-maximization. We empirically characterize the performance of our models on six text classification datasets. The choice of where to include the latent variable has a significant impact on performance, with the strongest results obtained when using the latent variable as an auxiliary conditioning variable in the generation of the textual input. This model consistently outperforms both the generative and discriminative classifiers in small-data settings. We analyze our model by using it for controlled generation, finding that the latent variable captures interpretable properties of the data, even with very small training sets. 
\end{abstract}
\section{Introduction}

The most widely-used neural network classifiers are \textbf{discriminative}, that is, they are trained to explicitly favor the gold standard label over others. The alternative is to design classifiers that are \textbf{generative}; these follow a generative story that includes predicting the label and then the data conditioned on the label. Discriminative classifiers are preferred because they generally outperform their generative counterparts on standard benchmarks. These benchmarks typically assume large annotated training sets, little mismatch between training and test distributions, relatively clean data, and a lack of adversarial examples~\citep{zue1990speech, marcus1993building, deng2009imagenet, lin2014microsoft}. 

However, when conditions are not ideal for discriminative classifiers, generative classifiers can actually perform better. \citet{ng2002discriminative} showed theoretically that linear generative classifiers approach their asymptotic error rates more rapidly than discriminative ones. Based on this finding, \citet{yogatama2017generative} empirically characterized the performance of RNN-based generative classifiers, showing advantages in sample complexity,  zero-shot learning, and continual learning. Recent work in generative question answering models~\citep{lewis2018generative} demonstrates better robustness to biased training data and adversarial testing data than state-of-the-art discriminative models.

In this paper, we focus on settings with small amounts of annotated data and improve generative text classifiers by introducing discrete latent variables into the generative story. Accordingly, the training objective is changed to log marginal likelihood of the data as we marginalize out the latent variables during learning. We parameterize the distributions with standard neural architectures used in conditional language models and include the latent variable by concatenating its embedding to the RNN hidden state before computing the softmax over words. While traditional latent variable learning in NLP uses the expectation-maximization (EM) algorithm~\citep{dempster1977maximum}, we instead simply perform direct optimization of the log marginal likelihood using gradient-based methods. At inference time, we similarly marginalize out the latent variables while maximizing over the label. 

We characterize the performance of our latent-variable generative classifiers on six text classification datasets introduced by \citet{Zhang:2015:CCN:2969239.2969312}. We observe that introducing latent variables leads to large and consistent performance gains in the small-data regime, though the benefits of adding latent variables reduce as the training set becomes larger.

To better understand the modeling space of latent variable classifiers, we explore several graphical model configurations. Our experimental results demonstrate the importance of including a direct dependency between the label and the input in the model. We study the relationship between the label, latent, and input variables in our strongest latent generative classifier, finding that the label and latent capture complementary information about the input. Some information about the textual input is encoded in the latent variable to help with generation. 

We analyze our latent generative model by generating samples when controlling the label and latent variables. Even with small training data, the samples capture the salient characteristics of the label space while also conforming to the values of the latent variable, some of which we find to be interpretable. While discriminative classifiers excel at separating examples according to labels, generative classifiers offer certain advantages in practical settings that benefit from a richer understanding of the data-generating process. 
\section{Discriminative and Generative Text Classifiers}
We begin by defining our baseline generative and discriminative text classifiers for document classification. Our models are essentially the same as those from \citet{yogatama2017generative}; we describe them in detail here because our latent-variable models will extend them.\footnote{The main difference between our baselines and the models in \citet{yogatama2017generative} are: (1) their discriminative classifier uses an LSTM with ``peephole connections''; (2) they evaluate a label-based generative classifier (``Independent LSTMs'') that uses a separate LSTM for each label. They also evaluate the model we described here, which they call ``Shared LSTMs''. Their Independent and Shared LSTMs perform similarly across training set sizes.} Our classifiers are trained on datasets $D$ of annotated documents. Each instance $\langle x, y\rangle \in D$ consists of a textual input $x = \{x_1, x_2, ..., x_T\}$, where $T$ is the length of the document, and a label $y \in \mathcal{Y}$.

The discriminative classifier is trained to maximize the conditional  probability of labels given documents: $\sum_{\langle x,y \rangle \in D} \log p(y \mid x)$. For our discriminative model, we encode a document $x$ using an LSTM~\citep{hochreiter1997long}, and use the average of the LSTM hidden states as the document representation. The classifier is built by adding a softmax layer on top of the LSTM state average to get a probability distribution over labels.

The generative classifier is trained to maximize the joint probability of documents and labels: $\sum_{\langle x,y \rangle \in D} \log p(x, y)$. The generative classifier uses the following factorization:
\begin{equation}
p(x, y) = p(x \mid y) p(y)
\end{equation}
We parameterize $\log p(x \mid y)$ as a conditional LSTM language model using the standard sequential factorization:
\begin{equation}
\log p(x \mid y) = \sum_{t=1}^{T} \log p(x_t \mid x_{<t}, y)
\end{equation}
We define a label embedding matrix $V_\mathcal{Y} \in \mathbb{R}^{d_1 \times |\mathcal{Y}|}$. To predict the next word $x_{t+1}$, we concatenate the LSTM hidden state $h_t$ with the label embedding $v_y$ (a column of $V_\mathcal{Y}$), and feed it to a softmax layer to get the probability distribution over the vocabulary.  More details about the factorization and parameterization are discussed in Section~\ref{latentgenerativemodels}.
The label prior $p(y)$ is acquired via maximum likelihood estimation and fixed during training of the remaining parameters.

At inference time, the prediction is made by maximizing $p(y\mid x)$ with respect to $y$ for the discriminative classifier and maximizing $p(x\mid~y) p(y)$ for the generative classifier. 
\section{Latent-Variable Generative Classifiers}
\label{latentgenerativemodels}

\begin{figure}[t]
\centering
\includegraphics[width=0.28\textwidth]{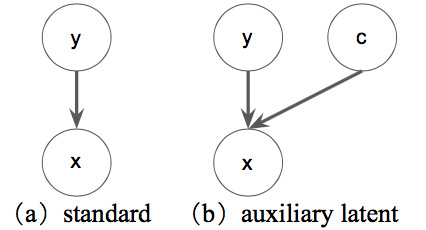}
\caption{Graphical models of (a) standard generative classifier and (b)  auxiliary latent generative classifier. 
\label{fig:aux_gen_graphic_model}}
\end{figure}

We now introduce discrete latent variables into the standard generative classifier as shown in Figure~\ref{fig:aux_gen_graphic_model}. We refer to the latent-variable model as an auxiliary latent generative model, as we expect the latent variable to contain auxiliary information that can help with the generation of the input.

Following the graphical model structure in Figure~\ref{fig:aux_gen_graphic_model}(b), we factorize the joint probability $p(x,y,c)$ as follows: 
\begin{align}
p(x,y,c) = p_{\Theta}(x \mid c, y) p_{\Phi}(c) p_{\Psi}(y)
\end{align}
We parameterize $p_{\Theta}(x \mid c, y)$ as a conditional LSTM language model using the same factorization as above: 
\begin{align}
\log p_{\Theta}(x \mid c, y) = \sum_{t=1}^{T} \log p_{\Theta}(x_t \mid x_{<t}, c, y)
\end{align}
where $\Theta$ is the set of parameters of the language model. As in the generative classifier, we use a label embedding matrix $V_\mathcal{Y}$. In addition, we define a latent variable embedding matrix $V_\mathcal{C} \in \mathbb{R}^{d_2 \times |\mathcal{C}|}$ where $\mathcal{C}$ is the set of values for the discrete latent variable. 
Also like the generative classifier, we use an LSTM to predict each word with a softmax layer: 
\begin{align}
 p_{\Theta}(x_t \mid x_{<t}, c, y) \propto \exp\{u_{x_t}^\top ([h_t; v_y; v_c]) + b_{x_t}\}
\end{align}
where $h_t$ is the hidden representation of $x_{<t}$ from the LSTM, $v_y$ and $v_c$ are the embeddings for the label and the latent variable, respectively, $[u;v]$ denotes vertical concatenation, $u_{x_t}$ is the output word embedding, and $b_{x_t}$ is a bias parameter. 

The prior distribution of the latent variable is parameterized as follows:
\begin{align}
p_{\Phi}(c) \propto \exp\{w_c ^\top v_c + b_c\}
\end{align}
where $\Phi$ is the set of parameters for this distribution which includes the vector $w_c$ and scalar $b_c$ for each $c$. 

As in the standard generative model, the label prior $p_{\Psi}(y)$ is acquired from the empirical label distribution in the training data and remains fixed during training. 

\paragraph{Training.}
As is standard in latent-variable modeling, we train our models by  maximizing the log marginal likelihood:

\begin{align}
\max_{\Theta, \Phi, V_\mathcal{C}, V_\mathcal{Y}}  \sum_{\langle x,y \rangle \in \mathcal{D}} \log \sum_{c \in \mathcal{C}} p_{\Theta}(x \mid c, y) p_{\Phi}(c) p_{\Psi}(y)
\end{align}
In NLP, these sorts of optimization problems are traditionally solved with the EM algorithm. However, we instead directly optimize the above quantity using automatic differentiation. This is natural because we use softmax-transformed parameterizations; a more traditional parameterization would assign parameters directly to individual probabilities, which then requires constrained optimization. 

\paragraph{Inference.}
\label{section:inference_marginalization}
The prediction is made by marginalizing out the latent variables as follows: 
\begin{align}
\hat{y} = \argmax_{y \in \mathcal{Y}} \sum_{c \in \mathcal{C}} p_{\Theta}(x \mid c, y) p_{\Phi}(c) p_{\Psi}(y)
\end{align}
\noindent We experimented with other inference objectives and found similar results. More details can be found in Appendix C.

\paragraph{Differences with ensembles.}
Our latent-variable model resembles an ensemble of multiple generative classifiers, but there are two main differences. First, all parameters in the latent generative classifier are trained jointly, while a standard ensemble combines predictions from multiple, independently-trained models. Joint training leads to complementary information being captured by latent variable values, as shown in our analysis. Moreover, a standard ensemble will lead to far more parameters (10, 30, or 50 times as many in our experimental setup) since each generative classifier is a completely separate model. Our approach simply conditions on the embedding of the latent variable value and therefore does not add many parameters.
\section{Experiments}

\subsection{Datasets}
We present our results on six publicly available text classification datasets introduced by \citet{Zhang:2015:CCN:2969239.2969312}, which include news categorization, sentiment analysis, question/answer topic classification, and article ontology classification.\footnote{A more detailed dataset description is in Appendix E.}

To compare classifiers across training set sizes, we follow the setup of \citet{yogatama2017generative} and construct multiple training sets by randomly sampling 5, 20, 100, 1k, 2k, 5k, and 10k instances per label from each dataset. 

\begin{figure*}[t]
\subfigure[Yelp Review Polarity]{
\label{Fig.sub.1}
\includegraphics[width=0.34\textwidth]{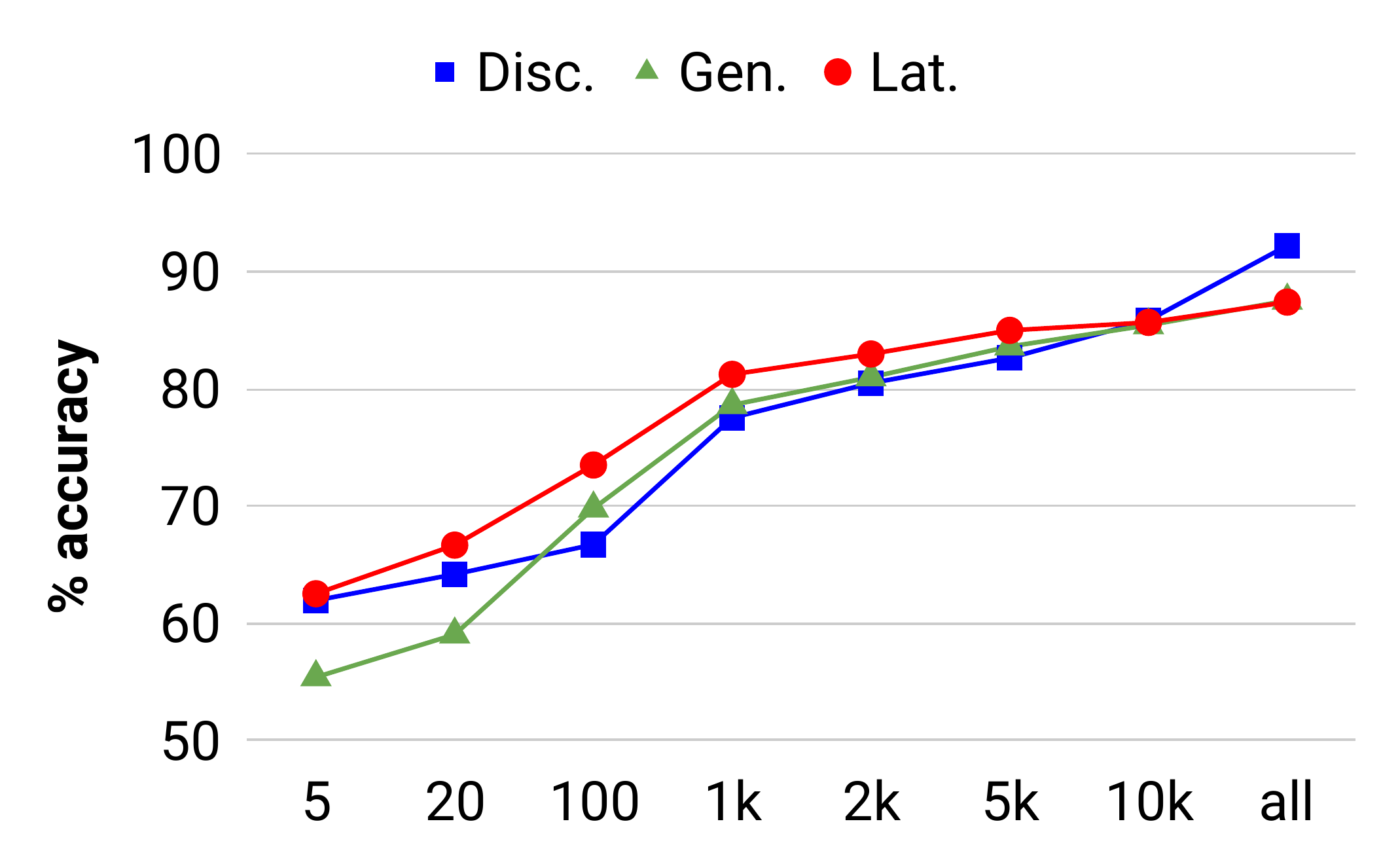}}
$\!\!\!\!\!\!\!\!$
\subfigure[Yelp Review Full]{
\label{Fig.sub.2}
\includegraphics[width=0.34\textwidth]{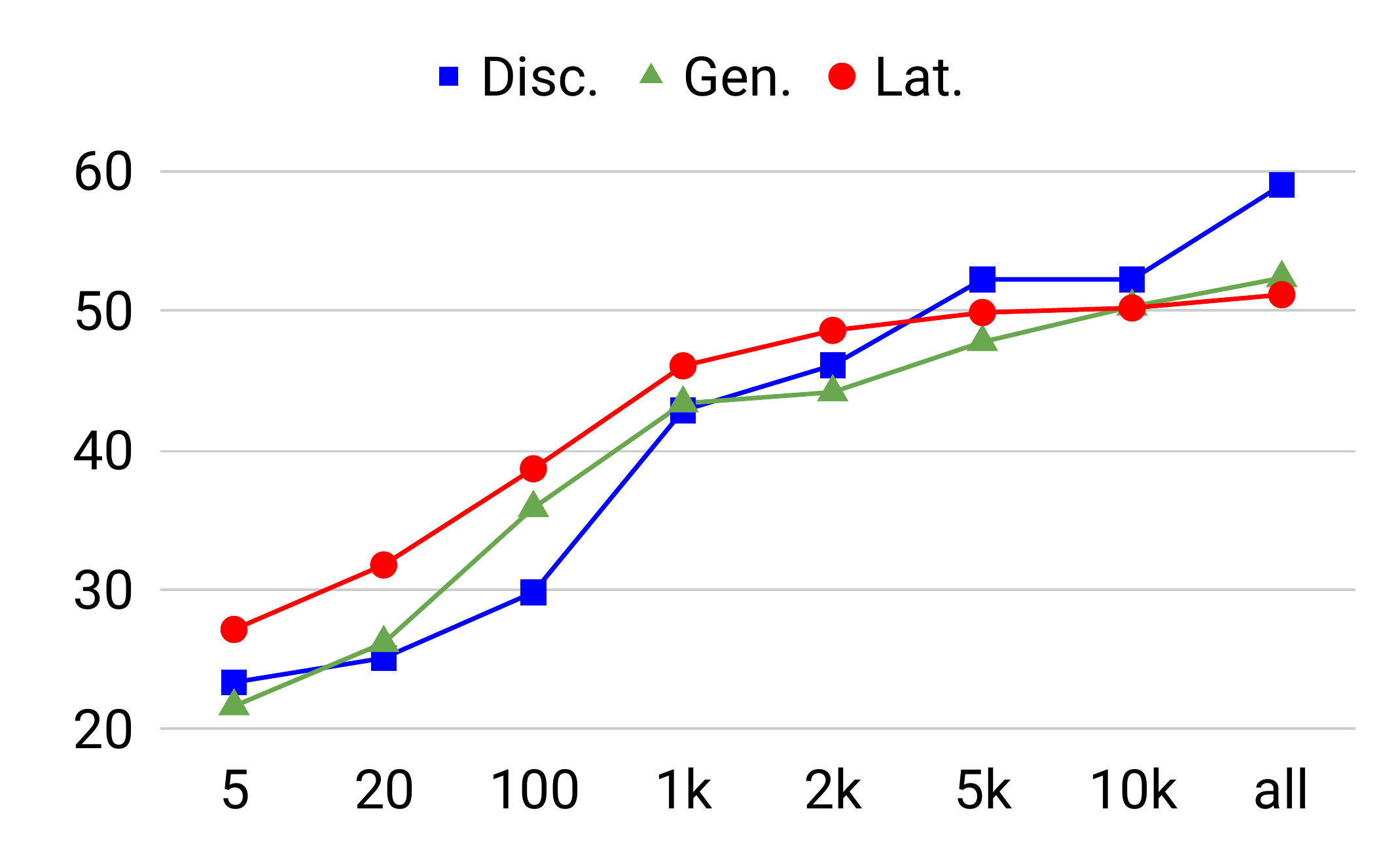}}
$\!\!\!\!\!\!\!\!$
\subfigure[AGNews]{
\label{Fig.sub.3}
\includegraphics[width=0.34\textwidth]{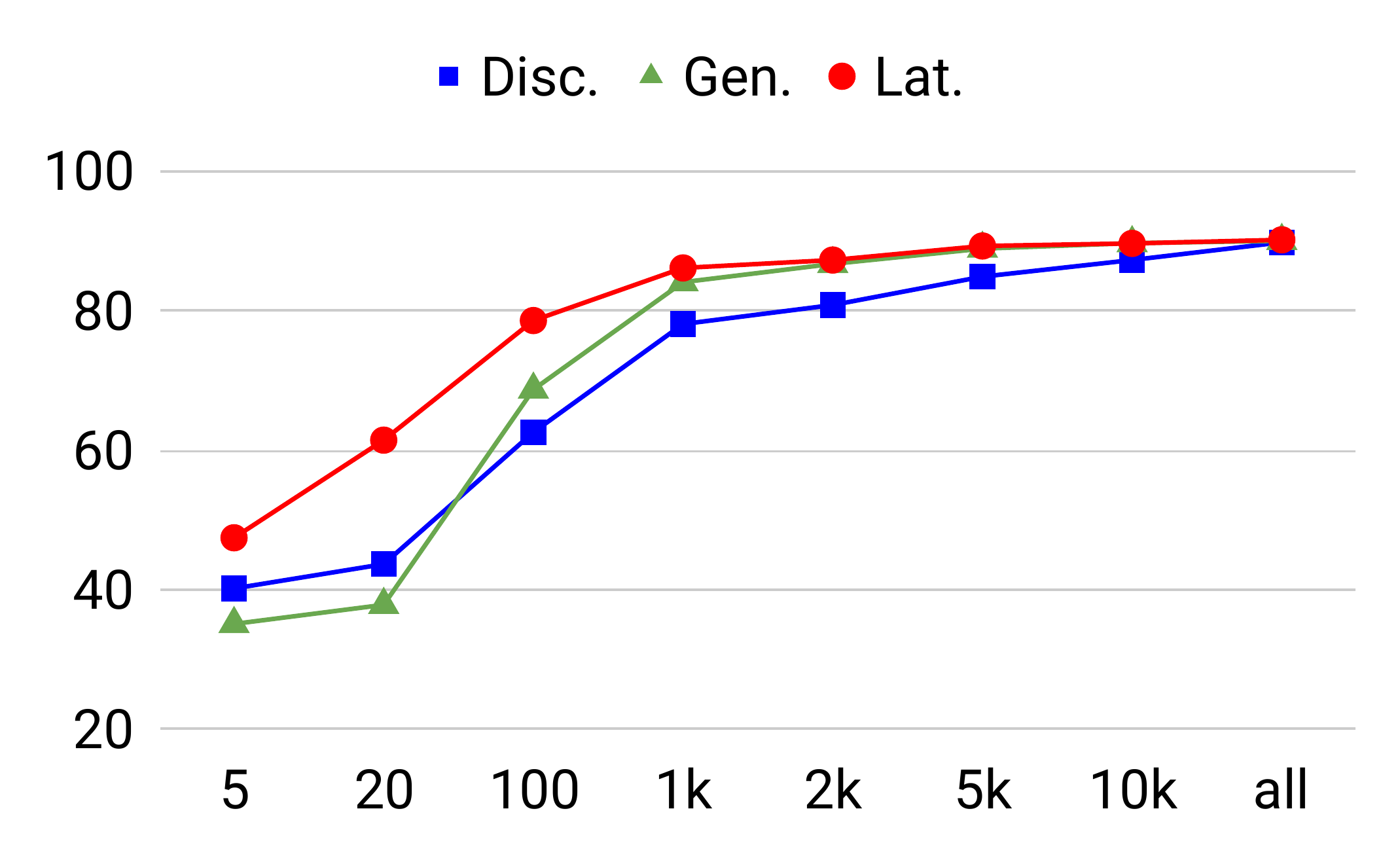}}\\
\subfigure[Sogou]{
\label{Fig.sub.4}
\includegraphics[width=0.34\textwidth]{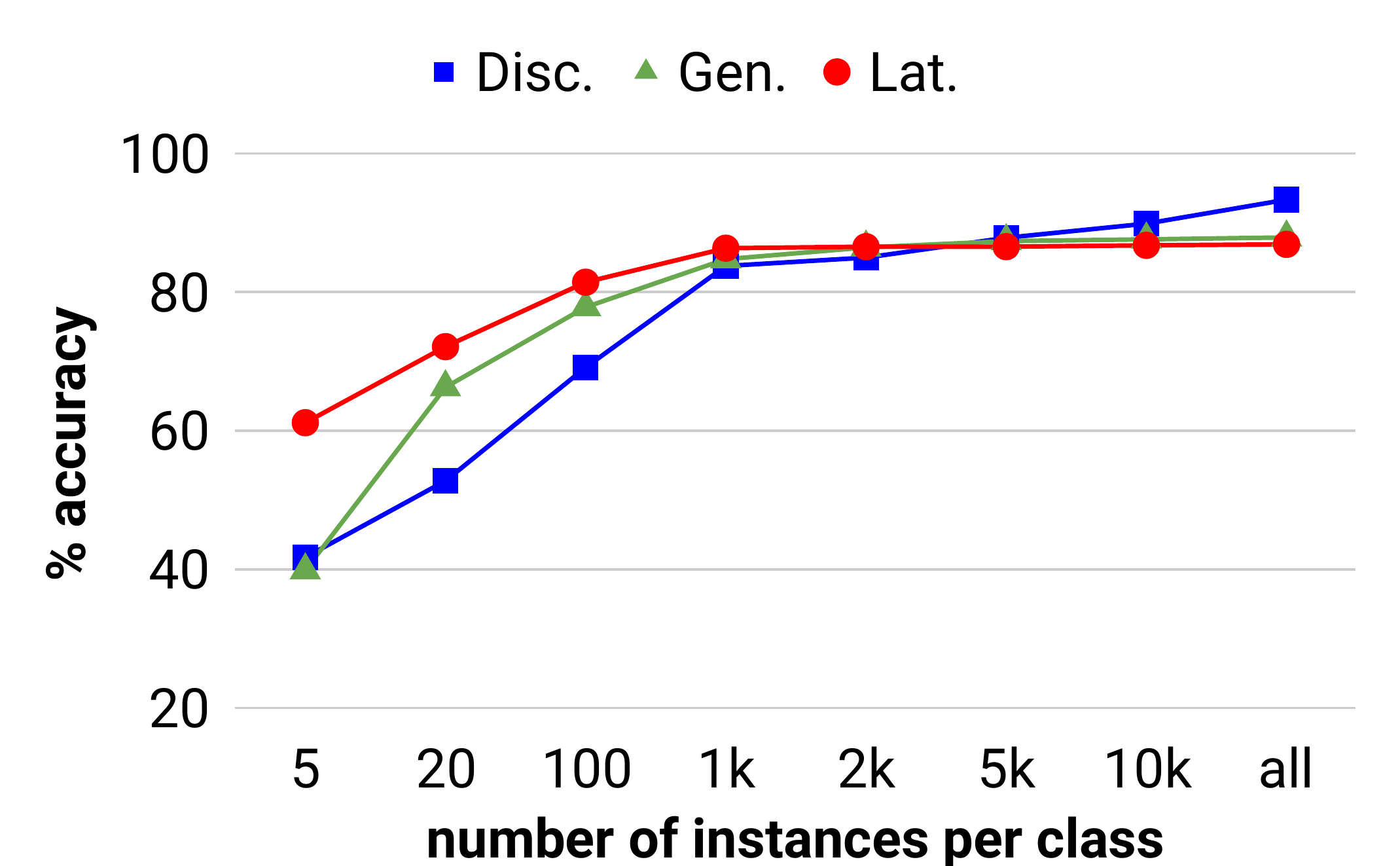}}
$\!\!\!\!\!\!\!\!$
\subfigure[Yahoo]{
\label{Fig.sub.5}
\includegraphics[width=0.34\textwidth]{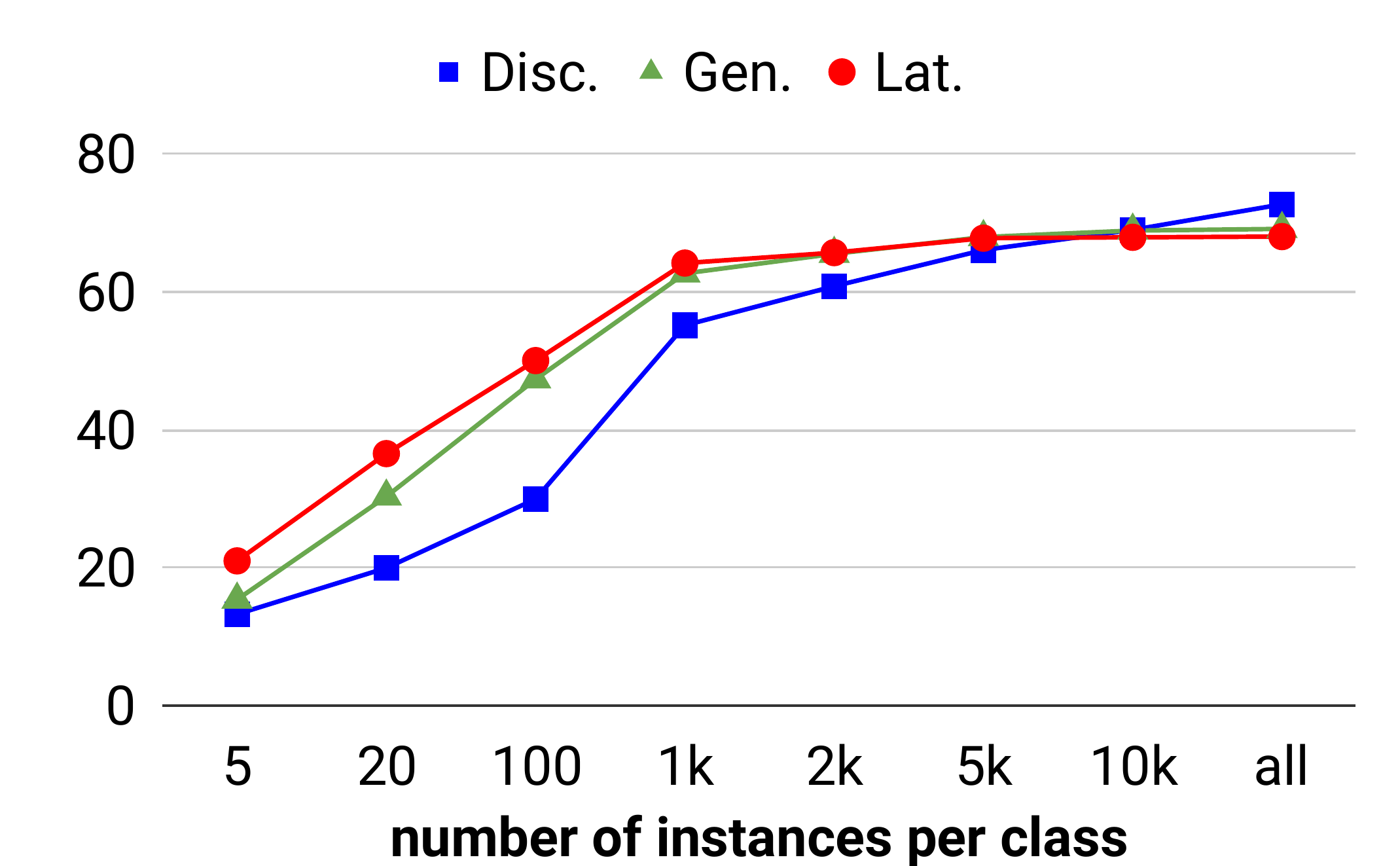}}
$\!\!\!\!\!\!\!\!$
\subfigure[DBpedia]{
\label{Fig.sub.6}
\includegraphics[width=0.34\textwidth]{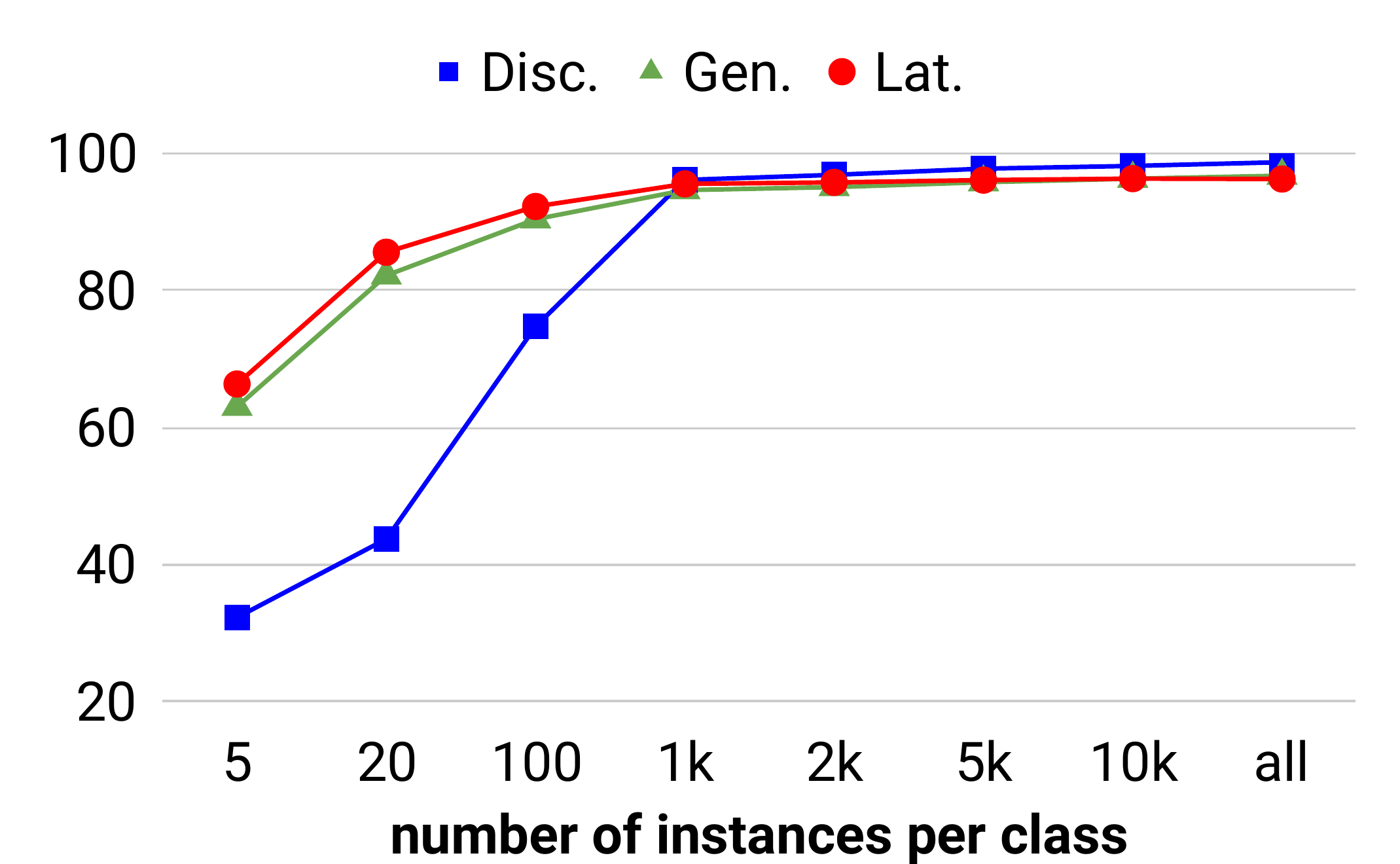}}
\caption{Comparison of classification accuracy of the discriminative (\textbf{Disc.}), standard generative (\textbf{Gen.}), and latent generative (\textbf{Lat.}) classifiers training across training set sizes. 
\label{fig:subset_compare}}
\end{figure*}

\subsection{Training Details}
\label{sec:details}
In all experiments, the word embedding dimension and the LSTM hidden state dimension are set to 100. All LSTMs use one layer and are unidirectional. The label dimensionality of all generative classifiers is set to 100. We adopt the same parameter settings as \citet{yogatama2017generative} to ensure the results are comparable. For the latent-variable generative classifiers, we choose 10 or 30 latent variable values with embeddings of dimensionality 10, 50, or 100.

For optimization, we use Adam~\cite{kingma2014adam} with learning rate 0.001. We do early stopping by evaluating the classification accuracy on the development set.

Due to memory limitations and computational costs, we truncate the length of the input sequences to 80 tokens before adding $\sos$ and $\eos$ to indicate the start and  end of the document. Though truncation decreases the performance of the models, all models use the same truncated inputs, so the comparison is still fair.\footnote{In other experiments, we compared performance with different truncation limits across training set sizes, finding the trends to be consistent with those presented here.} 

\subsection{Baselines}
To confirm we have built strong baselines, we first compare our implementation of the generative and discriminative classifiers to prior work. Our results in Appendix A show that our baselines are comparable to those of \citet{yogatama2017generative}.

\section{Results}
\subsection{Data Efficiency}

Figure \ref{fig:subset_compare} shows results for the discriminative, generative, and latent generative classifiers in terms of data efficiency. Data efficiency is measured by comparing the accuracies of the classifiers when trained across varying sizes of training sets. 
Numerical comparisons on two datasets are shown in Table~\ref{table:deltaAcc}.

With small training sets, the latent generative classifier consistently outperforms both the generative and discriminative classifiers. When the generative classifier is better than the discriminative one, as in DBpedia, the latent classifier resembles the generative classifier. When the discriminative classifier is better, as in Yelp Polarity, the latent classifier patterns after the discriminative classifier. However, when the number of training examples is in the range of approximately 5,000 to 10,000 per class, the discriminative classifier tends to perform best. 

With small training sets, the generative classifier outperforms the discriminative one in most cases except the very smallest training sets. For example, in the Yelp Review Polarity dataset, the first two points are from classifiers trained with only 10 and 40 instances in total. The other case in which generative classifiers underperform is when training over large training sets, which agrees with the theoretical and empirical findings in prior work~\citep{yogatama2017generative,ng2002discriminative}.

\begin{table}[t]
\begin{center}
\small
\begin{tabular}{|c|cc|cc|}
\cline{2-5}
\multicolumn{1}{c|}{}
&\multicolumn{2}{c|}{\textbf{$\Delta$(Lat., Gen.)}} & \multicolumn{2}{c|}{\textbf{$\Delta$(Lat., Disc.)}} \\ %\cline{2-5} 
\multicolumn{1}{c|}{}
                                                                                 & \textbf{AGNews}         & \textbf{DBpedia}         & \textbf{AGNews}         & \textbf{DBpedia}         \\ \hline
\textbf{5}                                                                                & +12.3                   & +3.3                   & +7.2                    & +34.0                 \\
\textbf{20}                                                                               & +23.5                   & +3.3                   & +17.7                   & +41.8                 \\
\textbf{100}                                                                              & +9.8                    & +1.8                   & +16.0                   & +17.5                  \\
\textbf{1k}                                                                             & +2.0                    & +0.9                   & +8.0                    & +0.0                  \\
\textbf{all}                                                                              & +0.1                    & -0.4                  & +0.3                   & -2.4                  \\ \hline
\end{tabular}
\end{center}
\caption{\textbf{$\Delta$(Lat.,~Gen.)}: change in accuracy when moving from generative to latent generative classifier; \textbf{$\Delta$(Lat.,~Disc.)}: change in accuracy when moving from discriminative to latent generative classifier. The first column shows the number of training instances per class.}
\label{table:deltaAcc}
\end{table}

\subsection{Effect of Graphical Model Structure}
\label{sec:latenteffect}
\begin{figure}[t]
\centering
\includegraphics[width=0.45\textwidth]{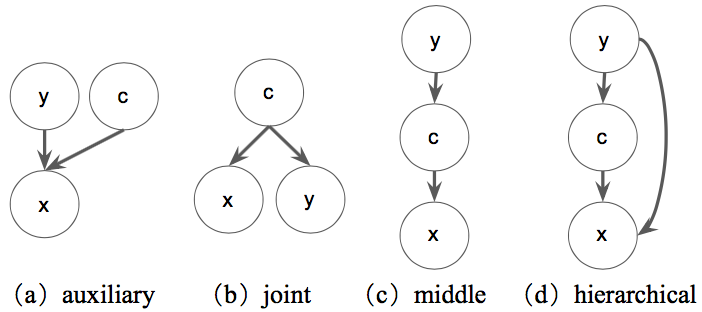}
\caption{Graphical models of (a) auxiliary, (b) joint, (c)  middle, and (d) hierarchical latent generative classifiers.
\label{fig:other_gen_graphic_model}}
\end{figure}

There are multiple choices to factorize the joint probability of the variables $x$, $y$, and $c$, which correspond to different graphical models. Here we consider other graphical model structures, namely those shown in Figure~\ref{fig:other_gen_graphic_model}. We refer to the model in Figure~\ref{fig:other_gen_graphic_model}(b) as the ``joint'' latent generative classifier since it uses the latent variable to jointly generate $x$ and $y$. We refer to the model in Figure~\ref{fig:other_gen_graphic_model}(c) as the ``middle'' latent generative classifier as the latent variable separates the textual input from the label. We use similar parameterizations for these models as for the auxiliary latent classifier, with conditional language models to generate $x$ where the embedding of the latent variable is concatenated to the hidden state as in Section~\ref{latentgenerativemodels}. 

\begin{figure*}[t]
\centering
\subfigure[Yelp Review Polarity]{
\label{gen_compare_2.sub1}
\includegraphics[width=0.34\textwidth]{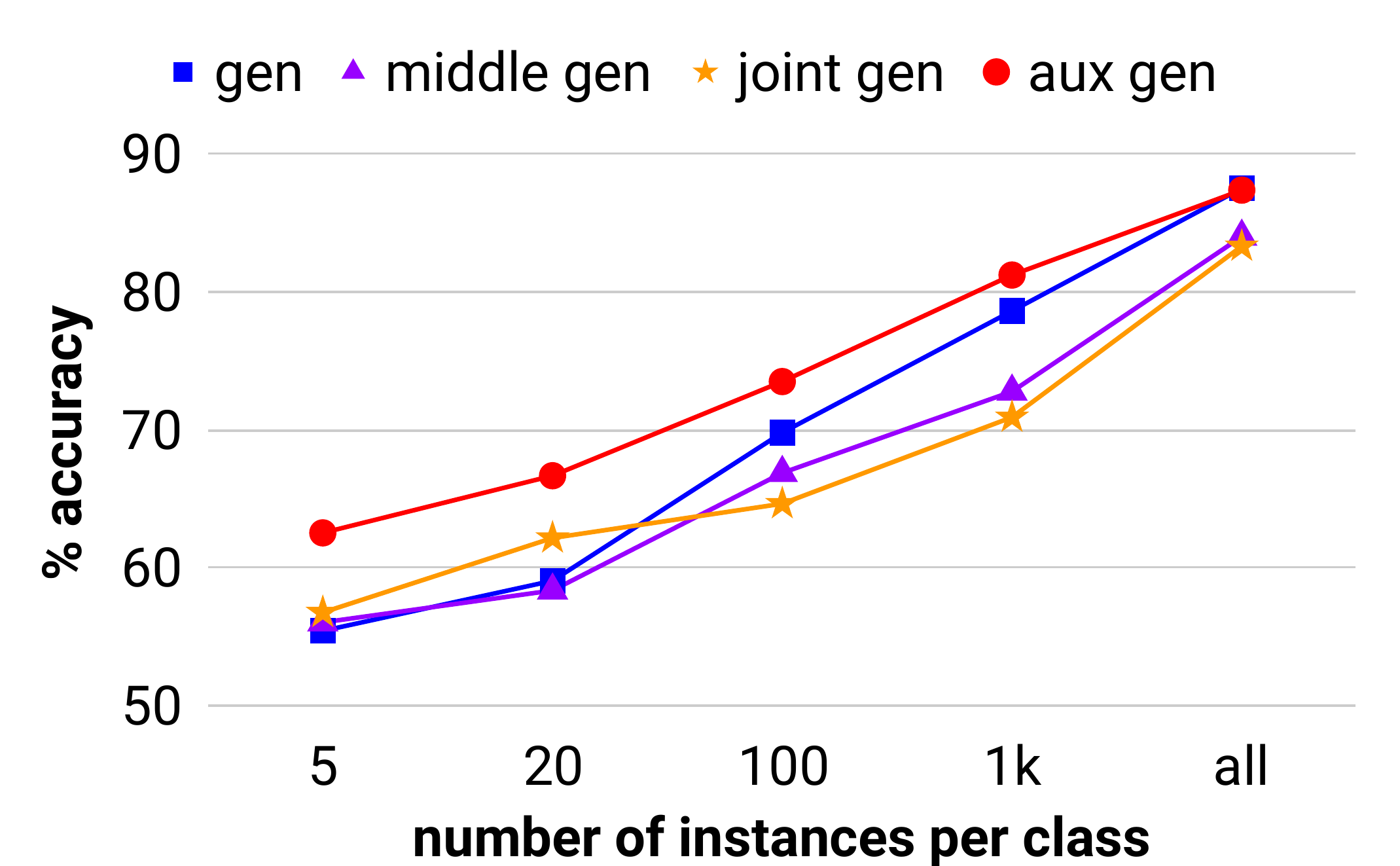}}
$\!\!\!\!\!\!\!\!$
\subfigure[AGNews]{
\label{gen_compare_2.sub2}
\includegraphics[width=0.34\textwidth]{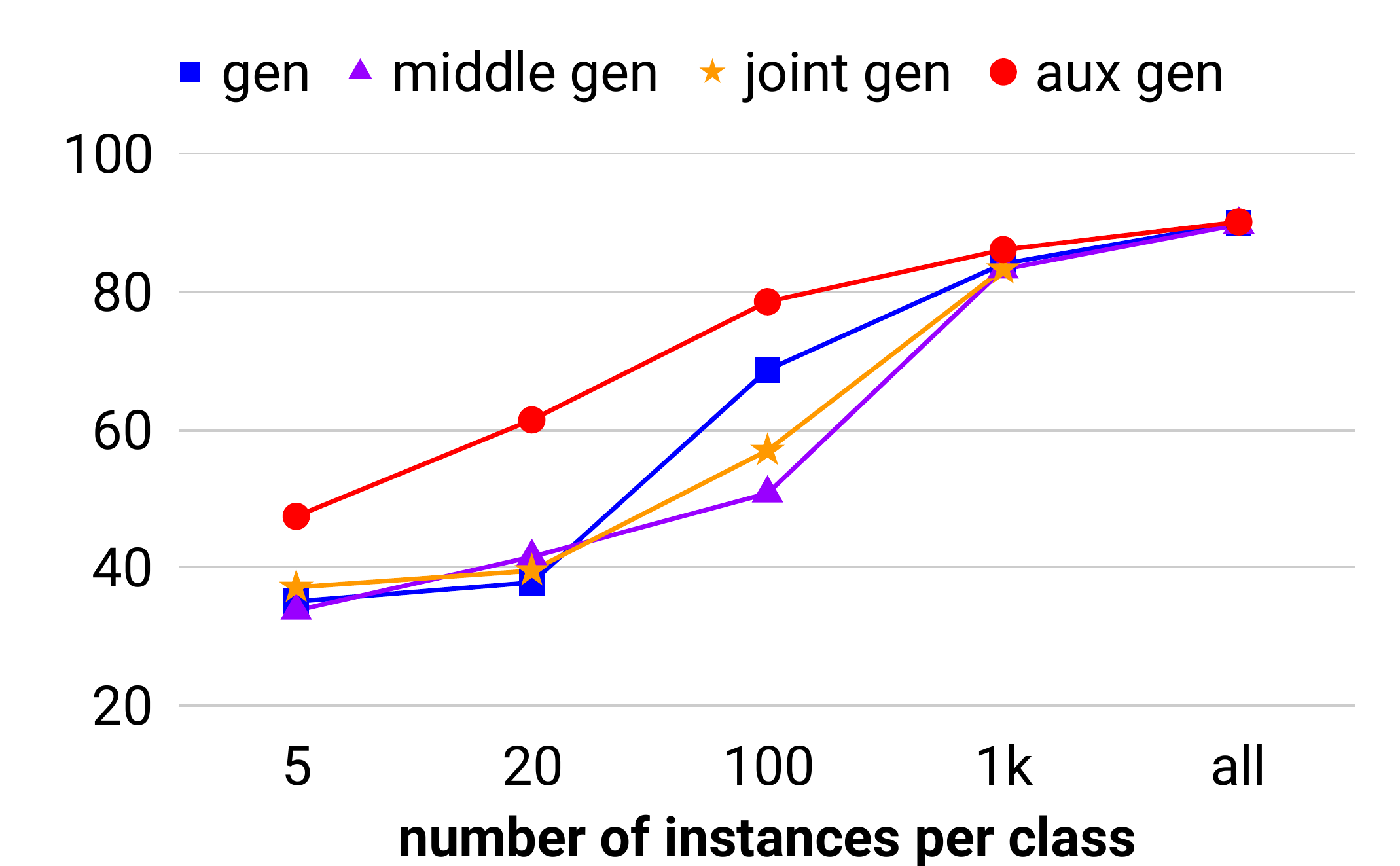}}
$\!\!\!\!\!\!\!\!$
\subfigure[DBpedia]{
\label{gen_compare_2.sub2}
\includegraphics[width=0.34\textwidth]{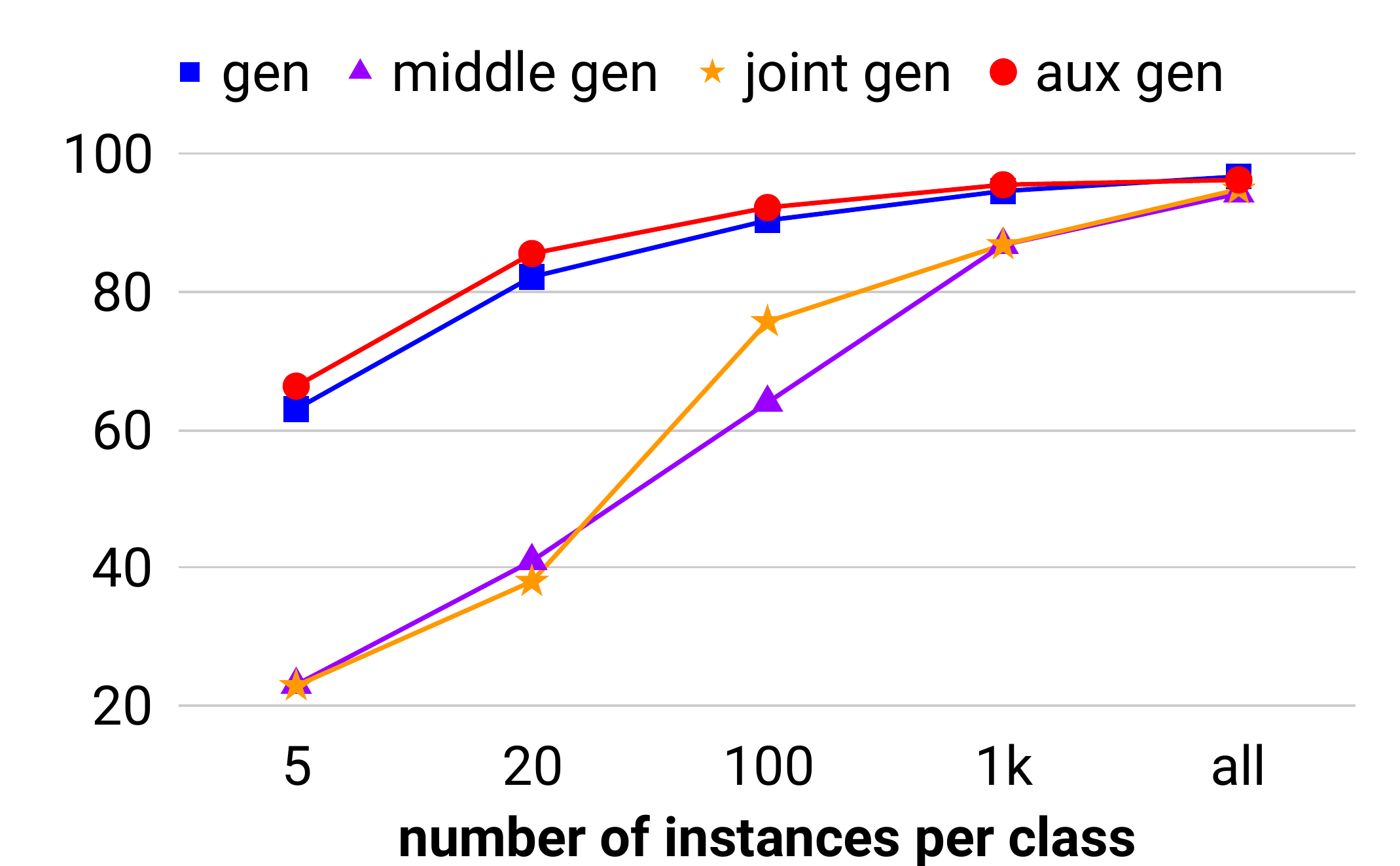}}
\caption{Comparison of generative classifier (\textbf{gen}) and latent generative classifiers (\textbf{middle gen, joint gen, aux gen}).
\label{fig:gen_compare_2}}
\end{figure*}

Figure \ref{fig:gen_compare_2} shows the comparison of the standard and the three latent generative classifiers on Yelp Review Polarity, AGNews, and DBpedia.\footnote{Similar trends are observed for all datasets, so we only show three for brevity.} We observe that the auxiliary model consistently performs best, while the other two latent generative classifiers do not consistently improve over the standard generative classifier. On DBpedia, we see surprisingly poor performance when adding latent variables suboptimally. This suggests that the choice of where to include latent variables has a significant impact on performance.

\paragraph{Dependency between label and input variable.}
We observe that the most prominent difference between the auxiliary and the other two latent-variable models is that the label variable $y$ is directly linked to the input variable $x$ in the auxiliary model, which is also the case in the standard generative model. 

\begin{table}[t]
\begin{center}
\small
\begin{tabular}{|c|c|c|c|c|c|}
\cline{2-6}
\multicolumn{1}{c|}{}
& \multicolumn{1}{c|}{\textbf{5}} & \multicolumn{1}{c|}{\textbf{20}} & \multicolumn{1}{c|}{\textbf{100}} & \multicolumn{1}{c|}{\textbf{1k}} & \multicolumn{1}{c|}{\textbf{all}} \\
\hline
\textbf{Yelp P} & +6.4       & +8.2        & +6.6         & +8.6          & +3.4         \\
\textbf{Yelp F} & +4.7      & +7.7        & +11.3        & +9.0          & +11.8        \\
\textbf{AGNews} & +13.4      & +19.9       & +27.9        & +1.8          & +0.4         \\
\textbf{Sogou}  & +22.3      & +24.0       & +13.9        & +3.3          & +2.8         \\
\textbf{Yahoo}  & +8.4       & +17.7       & +22.5        & +11.1         & +3.6         \\
\textbf{DBpedia}  & +44.6      & +44.6       & +28.4        & +8.8          & +2.3         \\ \hline
\end{tabular}
\end{center}
\caption{Changes in accuracy when adding a directed edge from the label to the input, i.e., the improvement in accuracy when moving from the middle to the hierarchical latent generative classifier. Each column shows a different number of training instances per class. }
\label{table:middle_hier_compare}
\end{table}

In order to verify the importance of this direct dependency between the label and input, we create a new latent-variable model by adding a directed edge between $y$ and $x$ to the middle latent generative model. 
We refer to this model as the ``hierarchical'' latent generative classifier, which is shown in Figure~\ref{fig:other_gen_graphic_model}(d). The results in Table \ref{table:middle_hier_compare} show the performance gains after adding this edge, which are all positive and sometimes very large. The resulting hierarchical model is very close in performance to the auxiliary model, which is unsurprising because these two models differ only in the presence of the edge from $y$ to $c$. 

\subsection{Effect of Latent Variables}
We conduct a comparison to demonstrate that the performance gains are due to the latent-variable structure instead of an increased number of parameters when adding the latent variables.\footnote{The results in the preceding sections use the models with  configurations tuned on the development sets. We follow the practice of \citet{yogatama2017generative} and fix label dimensionality to 100, as described in Section~\ref{sec:details}. The only tuned hyperparameters are the number of latent variable values and the dimensions of their embeddings.} 

\begin{table}[t]
\setlength{\tabcolsep}{5pt}
\begin{center}
\small
\begin{tabular}{|l|l|lllll|}
\cline{3-7}
                                \multicolumn{1}{c}{} & \multicolumn{1}{c}{}                         & \multicolumn{1}{|c}{\textbf{5}} & \multicolumn{1}{c}{\textbf{20}} & \multicolumn{1}{c}{\textbf{100}} & \multicolumn{1}{c}{\textbf{1k}} & \multicolumn{1}{c|}{\textbf{all}}  \\ \hline
\multirow{4}{*}{\textbf{Yelp P}} & \textbf{Gen.}            & 55.4 & 59.0 & 69.8 & 78.6 & 87.5 \\  
                                 & \textbf{Gen.~PC}     & 55.4 & 58.4 & 69.5 & 78.5 & 87.5 \\  
                                 & \textbf{Lat.}        & 62.5 & 66.6 & 73.5 & 81.2 & 87.3 \\  
                                 & \textbf{Lat.~PC} & 62.2 & 66.2 & 73.0 & 80.8 & 87.1 \\ \hline
\multirow{4}{*}{\textbf{AGNews}} & \textbf{Gen.}            & 35.1 & 37.8 & 68.7 & 84.0 & 90.0 \\  
                                 & \textbf{Gen.~PC}     & 33.7 & 37.8 & 68.4 & 83.2 & 89.8 \\  
                                 & \textbf{Lat.}        & 47.4 & 61.4 & 78.5 & 86.1 & 90.1 \\  
                                 & \textbf{Lat.~PC} & 47.2 & 61.4 & 78.3 & 84.5 & 90.1 \\ \hline
\end{tabular}
\end{center}
\caption{Accuracy comparison of standard generative (\textbf{Gen.}) and latent (\textbf{Lat.}) classifiers under earlier experimental configurations and parameter-comparison configurations (\textbf{PC}). When controlling for the number of parameters, the latent classifier still outperforms the standard generative classifier, which indicates the performance gains are due to the latent variables instead of an increased number of parameters.
}
\label{table:latent_var_config}
\end{table}

For the latent generative classifier, we choose 10 latent variable values with embeddings of dimensionality 10, and a label dimensionality of 100 (\textbf{Lat.~PC} in Table~\ref{table:latent_var_config}). For the standard generative classifier, we choose a label dimensionality of 110 (\textbf{Gen.~PC} in Table~\ref{table:latent_var_config}). So, the numbers of parameters are comparable, since we ensure the same number of parameters in the ``output'' word embeddings in the softmax layer of the language model, which is the decision that most strongly affects the number of parameters.

Table~\ref{table:latent_var_config} shows the results with different configurations, including the choices mentioned above as well as the results from earlier configurations mentioned in the paper. We observe that the latent generative classifiers still perform better in terms of data efficiency, which shows that the latent-variable structure accounts for the performance gains. 

\subsection{Learning via Expectation-Maximization}
The results reported before are evaluated on the classifiers trained by directly maximizing the log marginal likelihood via gradient-based optimization. In addition, we train our latent generative classifiers with the EM algorithm~\citep{salakhutdinov2003relationship}. More training details can be found in Appendix B.

\begin{table}[t]
\begin{center}
\small
\begin{tabular}{|c|cc|cc|}
\cline{2-5}
\multicolumn{1}{c|}{}
& \multicolumn{2}{c|}{\textbf{AGNews}} & \multicolumn{2}{c|}{\textbf{Sogou}} \\ 
\multicolumn{1}{c|}{} & \textbf{Direct}      & \textbf{EM}      & \textbf{Direct}      & \textbf{EM}     \\ \hline
\textbf{5}                 & 47.4 (62)        & 47.4 (62)       & 61.1 (144)       & 61.2 (147)     \\
\textbf{20}                & 61.4 (72)        & 61.1 (69)       & 72.1 (30)        & 72.2 (29)      \\
\textbf{100}               & 78.5 (10)        & 78.5 (10)       & 81.4 (11)        & 81.4 (11)      \\
\textbf{1k}              & 86.1 (18)        & 86.1 (18)       & 86.4 (9)         & 85.9 (9)       \\
\textbf{all}               & 90.1 (7)         & 90.0 (5)        & 86.9 (0)         & 86.9 (0)       \\ \hline
\end{tabular}
\end{center}
\caption{Comparison of the classification accuracy and convergence speed of the classifiers trained with direct optimization (\textbf{Direct}) of the log marginal likelihood and the EM algorithm (\textbf{EM}). The numbers inside the parentheses are the numbers of epochs required to reach the classification accuracies listed outside the parentheses.}
\label{table:EM}
\end{table}

To speed convergence, we use a mini-batch version of EM, updating the parameters after each mini-batch. Our results in Table~\ref{table:EM} show that the direct approach and the EM algorithm have similar performance in terms of classification accuracy and convergence speed in optimizing the parameters of our latent models. Similar trends appear for the other datasets. 
\section{Analysis}

\subsection{Interpretation of Latent Variables}
We take the strongest latent-variable model, the auxiliary latent generative classifier, and analyze the relationship among the latent, input, and label variables. We use the AGNews dataset, which contains 4 categories: world, sports, business, and sci/tech. The classifier we analyze has 10 values for the latent variable and is trained on a training set containing 1k instances per class. 

We first investigate the relationship between the latent variable and the label by counting co-occurrences. For each instance in the development set, we calculate the posterior probability distribution over the latent variable, and pick the value with the highest probability as the preferred latent variable value for that instance. This is reasonable since in our trained model, the posterior distribution over latent variable values is peaked. Then we categorize the data by their preferred latent variable values and count the gold standard labels in each group. We observe that the labels are nearly uniformly distributed in each latent variable value, suggesting that the latent variables are not obviously being used to encode information about the label. 

\begin{table*}[t!]
\begin{center}
\small
\begin{tabular}{|l|l|l|}
\hline
id & description                     & examples          \\ \hline
1  & future/progressive tenses             & \begin{tabular}[c]{@{}l@{}}... Commission \textbf{is likely to} follow opinion in the U.S. on the merger suit ...\\... to increase computer software exports \textbf{is beginning to} show results ...\end{tabular}                                                                                                                                                             \\ \hline
2  & past/perfect tense               & \begin{tabular}[c]{@{}l@{}}A screensaver targeting spam-related websites appears to \textbf{have been} too successful .\\ Universal \textbf{has signed} a handful of artists to a digital-only record label . ..\end{tabular}                                                                                                                            \\ \hline
3  & region names, locations & \begin{tabular}[c]{@{}l@{}}\textbf{Newcastle} manager Bobby Robson ... relieved of his duties ... \textbf{Newcastle} announced ...\\ \textbf{ABUJA} ... its militias \textbf{in Darfur} before they would sign ...\end{tabular}                                                                                                     \\ \hline
4  & mixture                 &                                                                                                                                                                                  \\ \hline
5  & abbreviations             & \begin{tabular}[c]{@{}l@{}}\textbf{St.} Louis advanced to the \textbf{N.L.} championship series for the third time in five years ...\\ \textbf{UAL ( UALAQ.OB : OTC BB} - news - research ) ... ( \textbf{UAIRQ.OB : OTC BB} ... \end{tabular}                                                 \\ \hline
6  & numbers, money-related   & \begin{tabular}[c]{@{}l@{}}...challenge larger rivals in the fast-growing \textbf{2.1 billion-a-year} sleep aid market .... \\ ... to a \textbf{\$ 25,000 prize} , and more importantly , into the history books ...\end{tabular}                                                                                                                   \\ \hline
7  & dates               & \begin{tabular}[c]{@{}l@{}}... an Egyptian diplomat said \textbf{on Friday}, and the abduction of ... \textbf{earlier this month} .\\ ... expected \textbf{Monday or Tuesday} , ... doctors and nurses off for the \textbf{holiday weekend} ...\end{tabular}                \\ \hline
8  & country-oriented terms            & \begin{tabular}[c]{@{}l@{}}\textbf{Rwandan} President ... in the \textbf{Democratic Republic of the Congo} after ... \\ Pope John Paul II issued a new appeal for peace \textbf{in Iraq and the Middle East} ...\end{tabular} \\ \hline
9  & mixure                   &                                                                                                                                                                        \\ \hline
10 & symbols, links           & \begin{tabular}[c]{@{}l@{}}... \textbf{A HREF = " http : / / www.reuters.co.uk / financeQuoteLookup.jhtml } ... \\ \textbf{\& lt ;} strong \textbf{\& gt ;} Analysis \textbf{\& lt ; /} strong \textbf{ \& gt ;} Contracting out the blame ...\end{tabular}  \\ \hline
\end{tabular}
\end{center}
\caption{Latent variable values (``id''), our manually-defined descriptions, and examples of instances associated to them. Boldface is used to highlight cues to our labeling.  We use the term ``mixture'' when we did not find clear signals to interpret the latent variable value. }
\label{table:latent_interpretation}
\end{table*}

Thus, we hypothesize there should be information other than that pertaining to the label that causes the data to cluster into different latent variable values. We study the differences of the input texts among the 10 clusters by counting frequent words, manually scanning through instances, and looking for high-level similarities and differences. We report our manual labeling for the latent variable values in Table~\ref{table:latent_interpretation}. 

For example, value 1 is mostly associated with future and progressive tenses; the words ``will'', ``next'', and ``new'' appear frequently. Value 2 tends to contain past and perfect verb tenses (the phrases ``has been'' and ``have been'' appear frequently). Value 3 contains region names like ``VANCOUVER'', ``LONDON'', and ``New Brunswick'', while value 7 contains country-oriented terms like ``Indian'', ``Russian'', ``North Korea'', and
``Ireland''. Our choice of only 10 latent variable values causes them to capture the coarse-grained patterns we observe here. It is possible that more fine-grained differences would appear with a larger number of values. 

\subsection{Generation with Latent Variables}
Another advantage of generative models is that they can be used to generate data in order to better understand what they have learned, especially in seeking to understand latent variables. 
We use our auxiliary latent generative classifier to generate multiple samples by setting the latent variable and the label. Instead of the soft mixture of discrete latent variable values that is used in classification (since we marginalize over the latent variable at test time), here we choose a single latent variable value when generating a textual sample.

\begin{table*}[t!]
\begin{center}
\small
\begin{tabular}{|p{1cm}|p{14cm}|}
\hline
\multicolumn{2}{|l|}{\textbf{Latent variable id = 3: region names, locations}}    \\ \hline
world     & \textbf{BEIJING} ( Reuters ) - \textbf{Oklahoma} supporters unemployment claims that he plans to trying to restore access next season 's truce by ruling , saying a major parliament .                                                                                           \\ \hline
sport    & The \textbf{Dallas} Cowboys today continued advantage today with \textbf{Miami} and the Hurricanes had to get the big rotation for the first time this year .                                                                                                                                                                                                  \\ \hline
business & \textbf{Las Vegas} took one more high-stepping kick across the pond as casino operator Caesars Entertainment Inc .                                                                                                                                                                                                                        \\ \hline
sci/tech & \textbf{SAN FRANCISCO} - Sun Microsystems on Monday will surely offer the deal to sell up pioneer members into two years and archiving .                                                                                                                                                                                                  \\ \hline
\multicolumn{2}{|l|}{\textbf{Latent variable id = 6: numbers, money-related}}                                                                                                                                                                                                                                                                                                                  \\ \hline
world     & An Israeli helicopter gunship fired a missile among \textbf{\$ 5} million in to Prime Minister Ariel Sharon on the streets of U.S. warming may not be short-lived .                                                                                           \\ \hline
sport    & On Wednesday it would win the disgruntled one of the season opener in a \# 36 ; \textbf{8.75 billion} of World Cup final day for second .                                                                                                                                                                                                  \\ \hline
business & Reuters - U.S. drug company Biogen Idec \ is considering an all-share bid of more than \textbf{8.5 billion euros} \ ( \# 36 ; \textbf{10.6 billion} ) for Irish peer Elan , a newspaper reported on Sunday .                                                                                                                                      \\ \hline
sci/tech & The JVC Everio GZ-MC100 ( \textbf{\$ 1199.95} ) and GZ-MC200 ( \textbf{\$ 1299.95} ) will use \textbf{4GB} Microdrive cards , which are removable hard drives measuring \textbf{1.5} inches square , but will also lost vital " fans to recently over .                                                                                                            \\ \hline   
\multicolumn{2}{|l|}{\textbf{Latent variable id = 10: symbols, links}}                                                                                                                                                                                                                                                                                                                  \\ \hline
world     & A German court is set to hear all its secular oil , and western Kerik in Fallujah . \textbf{\& lt ; A HREF = " http :   / / www.investor.reuters.com / FullQuote.aspx ? ticker = Agency target = \/ Army} ...                                                                                           \\ \hline
sport    & White Sox to an overpowering 49-0 victory over The world championship game . \textbf{\& lt ; br \& gt ; \& lt ; br \& gt ;} Comcast SportsNet                                                                                                                                                                                                 \\ \hline
business & NEW YORK ( Reuters ) - U.S. stocks climbed on Monday , with a steep decline in commodity prices and lower crude oil dented shares of Alcoa Inc . \textbf{\& lt ; A HREF = " http : / / www.investor.reuters.com / FullQuote.aspx ? ticker = GDT.N target = / stocks / quickinfo / fullquote " \& gt ; \& lt ; / A \& gt ;}.                                                                                                                                      \\ \hline
sci/tech & Spyware problems introduced a radio frequency code Thursday . \textbf{\& lt ; FONT face = " verdana , MS Sans Serif , arial , helvetica " size = " -2 " color = " \# 666666 " \& gt ; \& lt ; B \& gt ; -washingtonpost.com \& lt ; / B \& gt ; \& lt}                                                                                                            \\ \hline  
\end{tabular}
\end{center}
\caption{Generated examples by controlling the latent variables and labels (world, sport, business, sci/tech) with our latent classifier trained on a small subset of the AGNews dataset.}
\label{table:latent_generation}
\end{table*}

To increase generation diversity, we use temperature-based sampling when choosing the next word, where higher temperature leads to higher variety and more noise. 
We set the temperature to 0.6. Note that the latent-variable model here is trained on only 4000 instances (1k for each label) from AGNews, so the generated samples do suffer from the small size of data used in training the language model. Table~\ref{table:latent_generation} shows some generated examples. We observe that different combinations of the latent variable and label lead to generations that comport with both the labels and our interpretations of the latent variable values. 

We speculate that the reason our generative classifiers perform well in the data-efficient setting is that they are better able to understand the data via language modeling rather than directly optimizing the classification objective.  Our generated samples testify to the ability of generative classifiers to model the underlying data distribution even with only 4000 instances.
\section{Related Work}
\paragraph{Supervised Generative Models.}
Generative models have traditionally been used in supervised settings for many NLP tasks, including naive Bayes and other models for text classification~\citep{Maron:1961:AIE:321075.321084,yogatama2017generative}, Markov models for sequence labeling~\citep{church-1988-stochastic,bikel1999algorithm,brants2000tnt,zhou-su-2002-named}, and probabilistic models for parsing~\citep{magerrnan-marcus-1991-pearl,black-etal-1993-towards,eisner-1996-three,collins1997three,dyer2016recurrent}.  
Recent work in generative models for question answering \citep{lewis2018generative} learns to generate questions instead of directly penalizing prediction errors, which encourages the model to better understand the input data. Our work is directly inspired by that of \citet{yogatama2017generative}, who build RNN-based generative text classifiers and show scenarios where they can be empirically useful. 

\paragraph{Text Classification.} 
Traditionally, linear classifiers~\citep{mccallum1998comparison, joachims1998text, fan2008liblinear} have been used for text classification. Recent work has scaled up text classification to larger datasets with models based on logistic regression~\citep{joulin-etal-2017-bag}, convolutional neural networks~\citep{kim2014convolutional,Zhang:2015:CCN:2969239.2969312,conneau2016very}, and recurrent neural networks~\citep{xiao2016efficient, yogatama2017generative}, the latter of which is most closely-related to our models. 

\paragraph{Latent-variable Models.} 
Latent variables have been widely used in both generative and discriminative models to learn rich structure from data~\citep{petrov2007improved,petrov2008discriminative,blunsom2008discriminative,yu2009learning, morency_modeling_2008}. Recent work in neural networks has shown that introducing latent variables leads to higher representational capacity~\citep{kingma2013auto,chung2015recurrent,burda2015importance,ji2016latent}. However, unlike variational autoencoders~\citep{kingma2014adam} and related work that use continuous latent variables, our model is more similar to recent efforts that combine neural architectures with discrete latent variables and end-to-end training~\citep[\emph{inter alia}]{ji2016latent,DBLP:conf/iclr/KimDHR17,kong17,chen-gimpel-2018-smaller,wiseman-etal-2018-learning}. 
\section{Discussion and Future Work}
%\subsection{Effect of Pretrained Embeddings}
An alternative solution to the small-data setting is to use language representations pretrained on large, unannotated datasets~\citep{mikolov2013distributed,pennington2014glove,devlin2018bert}. 
In other experiments not reported here, we found that using pretrained word embeddings leads to larger performance improvements for the discriminative classifiers than the generative ones. 

Future work will explore the performance of latent generative classifiers in other challenging experimental conditions, including testing robustness to data shift and adversarial examples as well as zero-shot learning. Another thread of future work is to explore the performance of discriminative models with latent variables, and investigate combining pretrained representations with both generative and discriminative classifiers. 
\section{Conclusion}
We focused in this paper on 
%focuses on the goal of achieving data efficiency in NLP and improves 
improving the data efficiency of generative text classifiers by introducing discrete latent variables into the generative story. Our experimental results demonstrate that, with small annotated training data, latent generative classifiers have larger and more stable performance gains over discriminative classifiers than their standard generative counterparts. Analysis reveals interpretable latent variable values and generated samples, even with very small training sets. 
%architectures and analyzed what the latent variable learned to capture via manual analysis of input clusters and generated textual sequences for particular combinations of labels and latent variable values. 
%generative classifiers by interpretation with clustering and generation using different combination of latent variable and label. Our results suggest the potential advantages of generative models. % Contrary to the widely belief that discriminative models are always preferred, there are multiple cases that generative models shine. Besides the data-efficiency we have addressed in our paper, 

%is worth to explore in the future work.

\section*{Acknowledgments}
We would like to thank Lingyu Gao, Qingming Tang, and Lifu Tu for helpful discussions, Michael Maire and Janos Simon for their useful feedback, the anonymous reviewers for their comments that improved this paper, and Google for a faculty research award to K.~Gimpel that partially supported this research.

\appendix

\bibliography{emnlp-ijcnlp-2019}
\bibliographystyle{acl_natbib}
\end{document}

\maketitle

\appendix

\section{Comparison of Generative and Discriminative Classifiers}
\begin{table*}[t]
\begin{center}
\small
\begin{tabular}{|l|c|c|c|c|c|c|}
\hline
models                            & Yelp P & Yelp F & AGNews & Sogou & Yahoo & DBpedia \\ \hline
generative classifier (ours, shared-LSTM)   & 92.61         & 57.36     & 89.88  & 89.57 & 68.87 & 96.46   \\
discriminative classifier (ours)        & 96.37         & 65.81     & 90.09  & 96.43 & 73.10 & 98.78   \\ \hline
gen LSTM-shared \citep{yogatama2017generative}     & 88.20         & 52.70     & 90.60  & 90.30 & 69.30 & 95.40   \\
gen LSTM-independent \citep{yogatama2017generative} & 90.00         & 51.90     & 90.70  & 93.50 & 70.50 & 94.80   \\
disc model  \citep{yogatama2017generative}            & 92.60         & 59.60     & 92.10  & 94.90 & 73.70 & 98.70   \\ \hline
bag of words   \cite{Zhang:2015:CCN:2969239.2969312}                   & 92.20         & 58.00     & 88.80  & 92.90 & 68.90 & 96.60   \\
fastText    \cite{joulin-etal-2017-bag}                      & 95.70         & 63.90     & 92.50  & 96.80 & 72.30 & 98.60   \\
char-CRNN   \cite{xiao2016efficient}                      & 94.50         & 61.80     & 91.40  & 95.20 & 71.70 & 98.60   \\
very deep CNN  \cite{conneau2016very}                   & 95.70         & 64.70     & 91.30  & 96.80 & 73.40 & 98.70   \\ \hline
\end{tabular}
\caption{Summary of the results on the full datasets. Our implementation of the generative model share parameters among classes.}
\label{table:alltrainingset}
\end{center}
\end{table*}

To indicate we have built a strong baseline, we first compare our implementation of the generative and discriminative classifiers to prior work. Note that here we use the whole training set without any truncation on the sequence length. 

Table \ref{table:alltrainingset} shows that our standard generative and discriminative LSTM models are comparable with \citet{yogatama2017generative}. All other well-performing models listed in the table are discriminatve models that use more complex modeling methods such as attention to boost performance. Since the focus of our paper is the impact of adding latent variables to generative models, we do not use more complex techniques when building our baselines.

Our results show that our standard generative and discriminative LSTM models are comparable with \citet{yogatama2017generative}. We also see that the generative models have lower classification accuracies with the full training set, which agrees with the findings in the prior work.

\section{More Details about Learning with Expectation-Maximization}
EM provides a general purpose local search algorithm for learning parameters in probabilistic models with latent variables, and it has been widely used in much prior work and has shown its efficacy in terms of convergence ~\citep{ruder2018discriminative,neal1998view, dempster1977maximum}. 

\citet{salakhutdinov2003relationship} theoretically study the close relationship between EM and direct optimization approaches with gradient-based methods. Here we empirically characterize the performance of our auxiliary latent generative classifiers with different training methods, namely EM and stochastic gradient descent (SGD)~\citep{bottou2010large} for direct optimization. 

For our latent generative classifiers, the Expectation (E) step computes the posterior distributions over the latent variable:
\begin{equation}
\hat{p}(c \mid x,y) \leftarrow \frac{p(x, y, c)}{\sum_{c' \in \mathcal{C}} p(x , y, c')} \nonumber
\end{equation}
The Maximization (M) step seeks to find new parameter estimates by maximizing the following: 
\begin{equation}
\sum_{\langle x,y \rangle \in \mathcal{D}} \sum_{c \in \mathcal{C}} \hat{p}(c \mid x,y) \log p(x, y, c) \nonumber
\end{equation}

\section{Alternative Inference Criteria} 
\label{section:inference_maximization}
The classification accuracies of the auxiliary latent generative model in the main text are based on predictions made while marginalizing out the latent variable. In addition, we experiment with two other inference objectives. 
One uses the posterior $p(c \mid x,y)$ instead of the learned prior $p_{\Phi}(c)$ during marginalization:
\begin{align}
\hat{y} = \argmax_{y \in \mathcal{Y}} \sum_{c \in \mathcal{C}} p_{\Theta}(x \mid c, y) p(c \mid x,y) p_{\Psi}(y)  \nonumber
\end{align}
\noindent The other way is to predict by maximizing the latent variable:
\begin{align}
\hat{y} = \argmax_{y \in \mathcal{Y}} \,\max_{c \in \mathcal{C}} \, p_{\Theta}(x \mid c, y) p_{\Phi}(c) p_{\Psi}(y)  \nonumber
\end{align}
\noindent 
We find very similar performance with all three inference criteria, which agrees with our observation that the classifiers learn peaked prior and posterior distributions over the discrete latent variables.

\section{Additional Results with Training Sizes}

\begin{table}[t]
\small
\begin{center}
\begin{tabular}{|c|c|c|c|}
\hline
\textbf{\# per class}   & \textbf{Disc.} & \textbf{Gen.} & \textbf{Lat.} \\\hline
5&	61.97&	55.42&	62.53 \\
20&	64.19&	59.06&	66.67 \\
100&	66.72&	69.80&	73.50 \\
1k&	77.53&	78.62&	81.22 \\
2k&	80.48&	80.98&	82.96 \\
5k&	82.62&	83.60&	84.97 \\
10k&	85.83&	85.41&	85.65 \\
all&	92.20&	87.51&	87.38 \\ \hline
\end{tabular}
\end{center}
\caption{Comparison of classification accuracy on Yelp Review Polarity dataset.}
\label{table:larger_dataset_yelppolarity}
\end{table}

\begin{table}[t]
\small
\begin{center}
\begin{tabular}{|c|c|c|c|}
\hline
\textbf{\# per class}& \textbf{Disc.} & \textbf{Gen.} & \textbf{Lat.} \\\hline
5&	23.38&	21.67&	27.16 \\
20&	25.12&	26.20&	31.78 \\
100&	29.82&	35.87&	38.67 \\
1k&	42.85&	43.36&	46.05 \\
2k&	46.09&	44.16&	48.58 \\
5k&	52.23&	47.75& 49.86 \\	
10k&	52.23&	50.30& 50.19 \\	
all&	59.00&	52.34&	51.14 \\ \hline
\end{tabular}
\end{center}
\caption{Comparison of classification accuracy on Yelp Review Full dataset.}
\label{table:larger_dataset_yelpfull}
\end{table}

\begin{table}[t]
\small
\begin{center}
\begin{tabular}{|c|c|c|c|}
\hline
\textbf{\# per class}   & \textbf{Disc.} & \textbf{Gen.} & \textbf{Lat.} \\\hline
5 & 40.20 & 35.12 & 47.46 \\
20&	43.68&	37.86&	61.45 \\
100&	62.58&	68.70&	78.58 \\
1k&	78.08&	84.08&	86.12 \\
2k&	80.80&	86.70&	87.25 \\
5k&	84.87&	88.88&	89.26 \\
10k&	87.25&	89.67&	89.63 \\
all&	89.79&	90.00&	90.14 \\ \hline
\end{tabular}
\end{center}
\caption{Comparison of classification accuracy on AG News dataset.}
\label{table:larger_dataset_agnews}
\end{table}

\begin{table}[t]
\small
\begin{center}
\begin{tabular}{|c|c|c|c|}
\hline
\textbf{\# per class}& \textbf{Disc.} & \textbf{Gen.} & \textbf{Lat.} \\\hline
5&	41.75&	39.89&	61.19 \\
20&	52.80&	66.32&	72.18 \\
100&	69.18&	77.88&	81.48 \\
1k&	83.83&	84.81&	86.40 \\
2k&	85.04&	86.50& 86.61 \\	
5k&	87.90&	87.42&	86.62 \\
10k&	89.94&	87.67&	86.81 \\
all&	93.40&	87.95&	86.95 \\ \hline
\end{tabular}
\end{center}
\caption{Comparison of classification accuracy on Sogou dataset.}
\label{table:larger_dataset_sogou}
\end{table}

\begin{table}[t]
\small
\begin{center}
\begin{tabular}{|c|c|c|c|}
\hline
\textbf{\# per class}& \textbf{Disc.} & \textbf{Gen.} & \textbf{Lat.} \\\hline
5&	13.26&	15.39&	21.00 \\
20&	19.98&	30.33&	36.55 \\
100&	29.97&	47.33&	50.04 \\
1k&	55.15&	62.68&	64.18 \\
2k&	60.83&	65.52&	65.70 \\
5k&	66.07&	67.95&	67.79 \\
10k&	69.00&	68.90&	67.92 \\
all&	72.70&	69.14&	68.02 \\ \hline
\end{tabular}
\end{center}
\caption{Comparison of classification accuracy on Yahoo dataset.}
\label{table:larger_dataset_yahoo}
\end{table}

\begin{table}[t]
\small
\begin{center}
\begin{tabular}{|c|c|c|c|}
\hline
\textbf{\# per class}& \textbf{Disc.} & \textbf{Gen.} & \textbf{Lat.} \\\hline
5&	32.27&	63.02&	66.33 \\
20&	43.72&	82.17&	85.56 \\
100&	74.73&	90.37&	92.24 \\
1k&	96.11&	94.62&	95.54 \\
2k&	96.85&	95.06&	95.75 \\
5k&	97.76&	95.78&	96.09 \\
10k&	98.15&	96.29&	96.30 \\
all&	98.70&	96.73&	96.25 \\ \hline
\end{tabular}
\end{center}
\caption{Comparison of classification accuracy on DBpedia dataset.}
\label{table:larger_dataset_dbpedia}
\end{table}

%\kevin{This is great to have all of these results. I think we should extend the plots in the main paper to include these results.} \xiaoan{Sure.}
While the main paper contained these results in plots, for completeness we provide the numerical classification accuracies of the discriminative (\textbf{Disc.}), generative (\textbf{Gen.}), and latent-variable generative (\textbf{Lat.})~classifiers trained with various training sizes in Tables \ref{table:larger_dataset_yelppolarity}, \ref{table:larger_dataset_yelpfull}, \ref{table:larger_dataset_agnews}, 
\ref{table:larger_dataset_sogou},
\ref{table:larger_dataset_yahoo}, and 
\ref{table:larger_dataset_dbpedia}. 
%The benefits of using generative models and adding latent variables reduce as the training set becomes larger, and eventually  underperform the discriminative counterparts. 

%Table \ref{table:small_large} shows numerical results comparing the discriminative and latent-variable classifiers with small and large training sets. 
%The results are consistent with the theoretically and empirically findings in prior work~\citep{yogatama2017generative, ng2002discriminative} that generative models reach their (higher) asymptotic error faster than the discriminative models. 
%Contrary to the widely belief that discriminative models always outperform, there are multiple cases that generative models are prefered. Besides the data-efficiency we have addressed in our paper, the performance of latent generative models in terms of robustness to data shift and adversarial examples and zero-shot learning are worth to explore further in the future work.

\section{Dataset Description}
We present our results on six publicly available text classification datasets introduced by \citet{Zhang:2015:CCN:2969239.2969312}, which 
%are classification tasks including 
include news categorization, sentiment analysis, question/answer topic classification, and article ontology classification. Dataset names and statistics are shown in Table \ref{table:dataset}. For each dataset, we randomly pick 5000 instances from the training set as the development set.

\begin{table}[t]
\begin{center}
\small
\begin{tabular}{|c|c|c|c|c|c|}
%\toprule
\hline
\textbf{Dataset} & \textbf{\#Train} & \textbf{\#Dev} & \textbf{\#Test} & \textbf{\#Labels} \\ 
\hline
%\midrule
Yelp Polarity    & 555k             & 5k             & 7.6k            & 2                  \\ %\hline
Yelp Full        & 645k             & 5k             & 50k             & 5                  \\ %\hline
AGNews           & 115k             & 5k             & 7.6k            & 4                  \\ %\hline
Sogou            & 445k             & 5k             & 60k             & 5                  \\ %\hline
Yahoo            & 1395k            & 5k             & 60k             & 10                 \\ %\hline
DBpedia          & 555k             & 5k             & 70k             & 14                 \\ 
\hline
\end{tabular}
\end{center}
\caption{Text classification datasets used in our experiments.}
\label{table:dataset}
\end{table}

% The data description is followed from the decriptions in \citet{conneau2016very, yogatama2017generative, Zhang:2015:CCN:2969239.2969312}

\section{Total Number of Parameters}

\begin{table*}[t]
\begin{center}
\small
\begin{tabular}{|c|c|c|c|c|c|}
\cline{2-6}
\multicolumn{1}{c|}{}
 & \textbf{word embedding} & \textbf{hidden state}  & \textbf{label embedding}  & \textbf{\# latent variable values}   & \textbf{latent embedding} \\
\hline
\textbf{Disc.} &	100&	100&	100&	-&	- \\
\textbf{Gen.} & 100&	100&	100&    -&	- \\
\textbf{Lat.} &	100&	100&	100&	10, 30, 50&	10, 50, 100 \\
\textbf{Gen. PC} &	100&	100&	110&	-&	- \\
\textbf{Lat. PC} &	100&	100&	100&	10&	10 \\ \hline
\end{tabular}
\end{center}
\caption{Hyperparameter settings of Discriminative (\textbf{Disc.}), Generative (\textbf{Gen.}), Latent Generative (\textbf{Lat.}), Generative PC (\textbf{Gen. PC}), Latent Generative PC (\textbf{Lat. PC}) classifiers. \textbf{PC} stands for  ``Parameter-comparison Configuration.'' More description can be found in the main paper. %\kevin{I think the ``10, 30, 50'' and ``10, 50, 100'' should be moved down one row, right?} \xiaoan{Thanks. Good Catch.}
}
\label{table:hyper-p}
\end{table*}

\begin{table*}[t]
\begin{center}
\small
\begin{tabular}{|c|c|c|c|c|c|c|}
\hline
\textbf{Dataset} & \textbf{\# per class} & \textbf{Disc.}  & \textbf{Gen.}  & \textbf{Lat.}   & \textbf{Gen.  PC}  & \textbf{Lat. PC} \\\hline
\multirow{5}{*}{Yelp P} & 5 & \multirow{5}{*}{4,082,414} & \multirow{5}{*}{12,642,828} &   16,122,903 &   \multirow{5}{*}{12,481,640}  &  \multirow{5}{*}{12,521,733}\\
    &   20  &   &   & 14,123,253 & &   \\
    &   100  &   &   & 14,123,253 & &   \\
    &   1k  &   &   & 14,123,253 & &   \\
    &   all  &   &   & 16,122,903 & &   \\ \hline

\multirow{5}{*}{Yelp F} & 5 & \multirow{5}{*}{4,082,414} & \multirow{5}{*}{12,642,828} &   12,522,233 &   \multirow{5}{*}{12,481,640}  &  \multirow{5}{*}{12,521,733}\\
    &   20  &   &   & 12,522,233 & &   \\
    &   100  &   &   & 12,522,233 & &   \\
    &   1k  &   &   & 12,522,233 & &   \\
    &   all  &   &   & 14,122,553 & &   \\ \hline

\multirow{5}{*}{AGNews} & 5 & \multirow{5}{*}{4,082,414} & \multirow{5}{*}{12,642,828} &   11,430,985 &   \multirow{5}{*}{10,104,780}  &  \multirow{5}{*}{10,137,185}\\
    &   20  &   &   & 10,137,185 & &   \\
    &   100  &   &   & 11,430,985 & &   \\
    &   1k  &   &   & 10,137,585 & &   \\
    &   all  &   &   & 12,522,333 & &   \\ \hline

\multirow{5}{*}{Sogou} & 5 & \multirow{5}{*}{4,082,414} & \multirow{5}{*}{12,642,828} &   13,162,568 &   \multirow{5}{*}{11,634,120}  &  \multirow{5}{*}{11,671,448}\\
    &   20  &   &   & 13,163,568 & &   \\
    &   100  &   &   & 15,028,468 & &   \\
    &   1k  &   &   & 11,676,935 & &   \\
    &   all  &   &   & 12,522,233 & &   \\ \hline

\multirow{5}{*}{Yahoo} & 5 & \multirow{5}{*}{4,082,414} & \multirow{5}{*}{12,642,828} &   12,522,533 &   \multirow{5}{*}{12,482,520}  &  \multirow{5}{*}{12,522,533}\\
    &   20  &   &   & 14,124,053 & &   \\
    &   100  &   &   & 12,522,733 & &   \\
    &   1k  &   &   & 12,522,533 & &   \\
    &   all  &   &   & 16,123,703 & &   \\ \hline

\multirow{5}{*}{DBpedia} & 5 & \multirow{5}{*}{4,082,414} & \multirow{5}{*}{12,642,828} &   12,522,933 &   \multirow{5}{*}{12,482,960}  &  \multirow{5}{*}{12,522,933}\\
    &   20  &   &   & 14,124,453 & &   \\
    &   100  &   &   & 14,124,453 & &   \\
    &   1k  &   &   & 16,126,103 & &   \\
    &   all  &   &   & 12,522,933 & &   \\ \hline
\end{tabular}
\end{center}
\caption{Number of parameters in each classifier.}
\label{table:parameters}
\end{table*}

Table~\ref{table:hyper-p} shows the hyperparameter settings for our classifiers. There are various choices of latent variable values and dimensionalities. We select the ones with the best classification accuracy according to the development sets. 

Table~\ref{table:parameters} lists the number of parameters in each classifier. It is related to the discussion about effect of latent structure in the main paper. We created \textbf{Gen. PC} and \textbf{Lat. PC} to demonstrate that the performance  gains are  due  to  the  latent-variable structure instead of an increased number of parameters when adding the latent variables.

\section{Results with Larger Models}

\begin{table*}[t]
\small
\begin{center}
\begin{tabular}{|c|c|c|c|c|c|c|c|}
\cline{2-8}
\multicolumn{1}{c|}{}   & \textbf{\# per class}& \textbf{Disc L.} & \textbf{Gen L.} & \textbf{Lat L.}& \textbf{Disc.} & \textbf{Gen.} & \textbf{Lat.} \\\hline
\multirow{5}{*}{AGNews} & 5& 37.34&	40.55&	47.91&	40.20&	35.12&	47.46 \\
 & 20& 44.53&	47.42&	62.36&	43.68&	37.86&	61.45 \\
 & 100& 62.74&	76.95&	79.63&	62.58&	68.70&	78.58 \\
 & 1k& 80.67&	84.82&	86.79&	78.08&	84.08&	86.12 \\
 & all& 90.54&	90.16&	89.68&	89.79&	90.00&	90.14 \\ \hline
 \multirow{5}{*}{Yelp Polarity} & 5& 60.82&	60.65&	63.07&	61.97&	55.42&	62.53 \\				
 & 20& 61.11&	64.44&	67.50&	64.19&	59.06&	66.67 \\
 & 100& 68.55&	71.79&	74.37&	66.72&	69.80&	73.50 \\
 & 1k& 77.93&	78.91&	81.82&	77.53&	78.62&	81.22 \\
 & all& 92.48&	87.76&	87.34&	92.20&	87.51&	87.38 \\ \hline
\end{tabular}
\end{center}
\caption{Comparision of classification accuracies between standard and larger classifiers.}
\label{table:larger_model}
\end{table*}

We increase the model capacity by increasing dimensionality of the word embedding, LSTM hidden embedding, and label embedding to 200 and refer to the resulting models as the large discriminative (\textbf{Disc L.}), generative (\textbf{Gen L.}), and latent-variable generative (\textbf{Lat L.}) classifiers. Note that we did not change the number of values or dimensionality of the latent variables. We only experimented with two datasets due to GPU memory limits.\footnote{The GPU memory consumption is affected by the number of labels in the generative and latent generative classifiers. These two datasets have relatively small numbers of labels.} Table \ref{table:larger_model} shows the performance comparison between standard (reported in the main paper) and larger classifiers. We find that the trend  \textbf{Disc~L.}~$<$~\textbf{Gen~L.}~$<$~\textbf{Lat~L.} still holds in most cases in the small-data setting, though the performance gaps shrink as the capacity increases.

\bibliography{emnlp-ijcnlp-2019}
\bibliographystyle{acl_natbib}